\documentclass[10pt,twocolumn,letterpaper]{article}

\usepackage{iccv}
\usepackage{times}
\usepackage{epsfig}
\usepackage{graphicx}
\usepackage{amsmath}
\usepackage{amssymb}

\usepackage{amsthm}
\usepackage{algorithm}
\usepackage{algorithmic}
\usepackage{multirow}
\usepackage{paralist}
\usepackage{wrapfig}

\def \R {\mathbb{R}}

\def \x {\mathbf{x}}
\def \OO {\mathcal{O}}

\def \cc {\mathbf{c}}

\def \w {\mathbf{w}}

\newtheorem{thm}{Theorem}
\newtheorem{lem}{Lemma}

\newcommand{\nop}[1]{}

\usepackage[pagebackref=true,breaklinks=true,letterpaper=true,colorlinks,bookmarks=false]{hyperref}

\iccvfinalcopy 

\def\iccvPaperID{9709} 

\ificcvfinal\fi

\begin{document}

\title{Weakly Supervised Representation Learning with Coarse Labels}

\author{Yuanhong Xu$^1$\quad Qi Qian$^2$\thanks{Corresponding author}\quad Hao Li$^1$\quad Rong Jin$^2$\quad Juhua Hu$^3$\\
$^{1}$ Alibaba Group, Hangzhou, China\\
$^{2}$ Alibaba Group, Bellevue, WA, 98004, USA\\
$^{3}$ School of Engineering and Technology\\ 
University of Washington, Tacoma, WA, 98402, USA\\
{\tt\small \{yuanhong.xuyh, qi.qian, lihao.lh, jinrong.jr\}@alibaba-inc.com, juhuah@uw.edu}
}

\maketitle
\ificcvfinal\thispagestyle{empty}\fi

\begin{abstract}
With the development of computational power and techniques for data collection, deep learning demonstrates a superior performance over most existing algorithms on visual benchmark data sets. Many efforts have been devoted to studying the mechanism of deep learning. One important observation is that deep learning can learn the discriminative patterns from raw materials directly in a task-dependent manner. Therefore, the representations obtained by deep learning outperform hand-crafted features significantly. However, for some real-world applications, it is too expensive to collect the task-specific labels, such as visual search in online shopping. Compared to the limited availability of these task-specific labels, their coarse-class labels are much more affordable, but representations learned from them can be suboptimal for the target task. To mitigate this challenge, we propose an algorithm to learn the fine-grained patterns for the target task, when only its coarse-class labels are available. More importantly, we provide a theoretical guarantee for this. Extensive experiments on real-world data sets demonstrate that the proposed method can significantly improve the performance of learned representations on the target task, when only coarse-class information is available for training. Code is available at \url{https://github.com/idstcv/CoIns}.
\end{abstract}

\section{Introduction}

Deep learning attracts more and more attentions due to its tremendous success in computer vision~\cite{GirshickDDM14,HeGDG17,KrizhevskySH12} and NLP applications~\cite{DevlinCLT19,MikolovSCCD13}. With modern neural networks, deep learning can even achieve a better performance than human beings on certain fundamental tasks~\cite{HeGDG17,SchroffKP15}. The improvement from deep learning makes many applications, e.g., autonomous driving~\cite{ChenMWLX17}, visual search~\cite{Qi2019}, question-answering system~\cite{YangHGDS16}, etc., become feasible.

\begin{figure}[ht]
\centering
\includegraphics[height=2.2in]{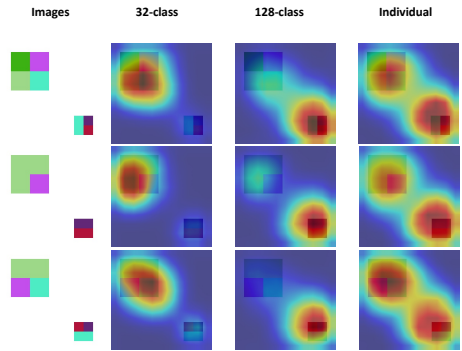}
\caption{Illustration of different patterns learned from different tasks on the same synthetic data consisting of $512$ images. According to different combinations of patches, three tasks are included: $32$ coarse-class classification (i.e., $32$-class with big patches), $128$-class classification (i.e., $128$-class with small patches) and instance-level classification (i.e., Individual with big and small patches). The detailed setting of the experiment can be found in appendix.\label{fig:2}}
\end{figure}


Compared with many existing models, which are designed for hand-crafted features, deep learning works in an end-to-end learning manner. It can explore the most discriminative patterns (i.e., features) from raw materials directly for a specific task. Without an explicit phase of generating features, deep learning demonstrates a significant improvement over existing methods~\cite{HeGDG17,KrizhevskySH12}. Using the features generated by deep learning, conventional methods can also perform better than the counterpart with hand-crafted features~\cite{Ting5709,DonahueJVHZTD14,GirshickDDM14, HeFWXG19}. This observation implies that neural networks can learn the task-related patterns sufficiently.

In deep learning, representations are often learned with respect to a specific task. Therefore, different patterns can be extracted even on the same data set for different application scenarios as shown in the example of Fig.~\ref{fig:2}. This phenomenon demonstrates that neural networks will only pay attention to those patterns that are helpful for the training task and ignore the unrelated patterns. Therefore, deep learning has to access a massive amount of labeled examples to achieve the ideal performance while the label information has to be closely related to the target task.

With the development of deep learning, a large training data size has been emphasized and many large-scale labeled data sets~\cite{DengDSLL009,LinMBHPRDZ14} become available. However, the correlation between the learned representations from provided labels and the target task is less investigated. In some real-world applications, it is often too expensive to gather task-specific labels, while their coarse-class labels are much more accessible. Taking visual search~\cite{Qi2019} as an example, given a query image of ``husky'', a result of ``husky'' is often expected than a ``dog''. Apparently, the label information like ``husky'' is much more expensive than that like ``dog''. The problem becomes more challenging in the online shopping scenario, where many items (e.g., clothes) have very subtle differences. The gap between the available labels and the target task makes the learned representations suboptimal.

To improve the performance of learned representations for a target task, a straightforward way is to label a sufficient number of examples specifically for that task, which can align the supervised information and the target task well. However, this strategy is not affordable. Unlike coarse-class labels, some task-specific labels (e.g., species of dogs) can only be identified by very experienced experts, which is expensive and inefficient. For the visual search task in the online shopping scenario, even experts cannot label massive examples accurately.

Recently, unsupervised methods become popular for representation learning~\cite{abs-2002-05709, DosovitskiyFSRB16, He0WXG20, abs-2105-11527, WuXYL18}. These methods first learn a deep model without any supervision on the source domain. After that, the learned model will be fine-tuned with the labeled data from the target domain. Although the pre-trained model is learned in an unsupervised manner, task-specific labels are required in the phase of fine-tuning, which are often very limited or have no access in some real-world applications. Considering that their coarse-class labels are much more affordable, in this work, we study the problem when data is from the target domain but only coarse-class labels are available.

Concretely, we aim to mitigate the issue by leveraging the information from coarse classes to learn appropriate representations for a target task. We verify that fine-grained patterns, which are essential for a target task, are often neglected when the deep model is trained only with coarse-class labels. Meanwhile, the popular pretext task in unsupervised representation learning, i.e., instance classification, may introduce too many noisy patterns that are irrelevant to the target task. Fortunately, we can theoretically prove that incorporating the task of coarse-class classification, representations learned from instance classification will be more appropriate for the target task. Based on this, we propose a new algorithm to learn appropriate representations for a target task when the task-specific labels are not available but their coarse-class labels are accessible. Besides, inspired by our analysis, a novel instance proxy loss is proposed to further improve the performance. Extensive experiments on benchmark data sets demonstrate that the proposed algorithm can significantly improve the performance on real-world applications when only coarse-class labels are available.

\section{Related Work}
\label{sec:related}

Different from many existing methods, deep learning can directly learn patterns from raw materials, which avoids the information loss in the phase of feature extraction. By investigating the patterns learned by deep neural networks, researchers find that it can adaptively figure out discriminative parts in images for classification due to the end-to-end learning manner~\cite{DonahueJVHZTD14,KrizhevskySH12}, which interprets the effectiveness of convolutional neural networks (CNNs). 

Besides supervised learning, unsupervised representation learning attracts much attention recently since it does not require any supervised information and can exploit the information from the large-scale unlabeled data sets~\cite{abs-2002-05709, DosovitskiyFSRB16, He0WXG20, WuXYL18}. A popular pretext task is instance classification~\cite{DosovitskiyFSRB16} that identifies each example as an individual class, while the computational cost can be a challenge on a large-scale data set. After its success, many algorithms are developed to improve the efficiency by contrastive learning~\cite{abs-2002-05709,He0WXG20, WuXYL18}. 

Despite the desired performance on the target domain, the process still relies on fine-tuning with labels of the target task. It is because that instance classification aims to identify each individual example and may introduce too many irrelevant patterns for the target task. Therefore, a fine-tuning phase is necessary to filter noisy patterns. Besides, the gap between the source and target domain may degrade the performance of learned representations~\cite{KornblithSL19,QianHL20}. In this work, we focus on the application scenario when the task-specific labels are hard to access, while their coarse-class labels (e.g., main categories for animals) are much cheaper and more accessible. We will leverage the weakly supervised information from coarse classes to improve the performance of learned representations on the target task, when target-specific labels are not available for fine-tuning.

It should be noted that generalizing learned models for different target tasks has also been researched in transfer learning and domain adaptation~\cite{ZhuangQDXZZXH21}. However, the problem addressed in this work is significantly different from them. Both of transfer learning and domain adaptation try to improve the performance on the target domain with the knowledge from a different source domain. In this work, we focus on learning with data from the target domain only.

\section{Proposed Method}
\label{sec:method} 

Given a set of $n$ images $\{(\x_i,y_i)\}_{i=1}^n$, a model can be learned by solving the optimization problem
\[\min_{\theta} \sum_{i=1}^n\ell(\x_i,y_i;\theta)\]
where $\ell(\cdot)$ is the loss function and $\theta$ denotes the parameters of a neural network. Cross-entropy loss with the softmax operator is a popular loss in deep learning. 

Many modern neural networks have multiple convolutional layers and a single fully-connected (FC) layer, e.g., ResNet~\cite{HeZRS16}, MobileNet~\cite{SandlerHZZC18}, EfficientNet~\cite{TanL19}, etc. We will investigate this popular architecture in this work, while the analysis can be extended to more generic structures. 

For a $K$-class classification problem, the cross-entropy loss can be written as
\[\ell(\x_i,y_i) = -\log\frac{\exp(f(\x_i)^\top \w_{y_i})}{\sum_j^K \exp(f(\x_i)^\top \w_j)}\]
where $f(\cdot)$ extracts features with convolutional layers from an image and $W=\{\w_1,\dots,\w_K\}\in\R^{d\times K}$ denotes the parameters of the last FC layer in a neural network. $d$ is the input dimension of FC layer when ignoring the bias term.

Apparently, the behavior of function $f$ heavily depends on the training labels in $\{y_i\}$. When the task implied by $\{y_i\}$ is consistent with the target one, the patterns discovered by $f$ can perform well. However, when the training task is different from the target one (e.g., 32-class labels for the 128-class target task in Fig.~\ref{fig:2}), the learned patterns can be suboptimal. In this work, we aim to learn an appropriate function $f$ that can extract sufficient and appropriate fine-grained patterns, even when only coarse-class labels are available.

\subsection{Instance Classification}
We start our analysis from the popular instance classification problem. The optimization problem for instance classification can be cast as
\begin{eqnarray}\label{eq:indi}
\min_{\theta} \sum_i \ell(\x_i, y_i^I;\theta)
\end{eqnarray}
where $y_i^I\in\{1,\dots,n\}$ and $y_i^I = i$. The problem in Eqn.~\ref{eq:indi} considers that each example is from a different class, which leads to an $n$-class classification problem. It can be more challenging than the classification problem with target labels and various patterns will be extracted to identify each individual example. However, the desired patterns for the target task can be overwhelmed by too many patterns obtained from instance classification. Therefore, the obtained representations can be far away from optimum, which is demonstrated in the following theoretical analysis.

Let $W^I\in\R^{d\times n}$ denote the parameters of the FC layer for instance classification. Both of $f^I$ and $W^I$ will be optimized as the parameters of the neural network. We define the prediction probability as 
\[\Pr\{y_i^I|f^I(\x_i), W^I\} = \frac{\exp(f^I(\x_i)^\top \w_{y^I_i}^I)}{\sum_j^n \exp(f^I(\x_i)^\top \w_j^I)}\]
It should be noted that we can have $\w_{y^I_i}^I = f^I(\tilde{\x}_i)$ in the contrastive learning~\cite{He0WXG20} where $\tilde{\x}_i$ is a different view of $\x_i$.
Assuming the task-specific labels are $y_i^F\in\{1,\dots, F\}$ and $F<n$, the performance of learned representations on the target task without fine-tuning can be evaluated by measuring the probability 
\[\Pr\{y_i^F|f^I(\x_i), W^I\} = \frac{\exp(f^I(\x_i)^\top \bar{\w}_{y_i^F}^I)}{\sum_s^F \exp(f^I(\x_i)^\top \bar{\w}_s^I)}\]
where $\bar{\w}_s^I= \frac{1}{z}\sum_{y_j^F = s} \w_j^I$. We assume each target class contains $z$ examples to simplify the analysis and $zF = n$. In this formulation, we adopt the mean vector of parameters from the same target class as the proxy for the target classification problem. The probability can measure the intra-class variance and inter-class distance in the learned representations. By investigating the performance, we can have the guarantee for the representations learned from instance classification as in the following Lemma. All detailed proofs of this work can be found in appendix.
\begin{lem}\label{lem:1}
If solving the problem in Eqn.~\ref{eq:indi} such that $\forall i, \Pr\{y_i^I|f^I(\x_i), W^I\}\geq \alpha$, we have
\begin{small}
\begin{align*}
&\forall i, \quad \Pr\{y_i^F|f^I(\x_i), W^I\}\geq z\alpha \exp(f^I(\x_i)^\top (\bar{\w}_{y_i^F}^I - \w_{y_i^I}^I))
\end{align*}
\end{small}
\end{lem}
\paragraph{Remark} Lemma~\ref{lem:1} shows that the performance of representations on the target task depends on both the accuracy of instance classification and the factor 
$f^I(\x_i)^\top \bar{\w}_{y_i^F}^I - f^I(\x_i)^\top\w_{y_i^I}^I$. When we have $\w_{y_i^I}^I =  f^I(\x_i)$ as in contrastive learning, the factor becomes $\frac{1}{z}\sum_{y_j^F=y_i^F} f^I(\x_i)^\top f^I(\x_j)$. Explicitly, the latter factor is corresponding to the intra-class variance.

Since instance classification is to identify every individual example, it can handle the inter-class difference well but the similarity between examples from the same target class can be arbitrary due to redundant patterns, which may result in a suboptimal performance. Therefore, we consider to leverage the coarse-class information to aggregate examples appropriately and filter irrelevant patterns to reduce intra-class variance.

\subsection{Intra-Class Optimization}
In many real-world applications, coarse-class labels (e.g., ``dog'', ``cat'', and ``bird'') are easy to access. The learning problem with coarse-class labels can be defined as
\begin{eqnarray}\label{eq:ori}
\min_{\theta} \sum_i \ell(\x_i, y_i^C;\theta)
\end{eqnarray}
where $y_i^C\in\{1,\dots, C\}$ indicates the coarse-class label of $\x_i$. In this work, we assume that examples from the same target class will share the same coarse labels. Explicitly, representations learned by solving this task can be inapplicable on a target task involving classes like ``bulldog'', ``husky'', and ``poodle'' under the coarse class ``dog''. It is because that the learned features have small intra-class variance but cannot handle the inter-class difference on the target classes. Consequently, they can separate the examples on the coarse classes well, while they cannot provide a meaningful separation for the target classes.

Based on these complementary observations from Eqns.~\ref{eq:indi} and~\ref{eq:ori}, we consider to incorporate the problem in Eqn.~\ref{eq:ori} to guide the learning of fine-grained patterns in Eqn.~\ref{eq:indi}. Intuitively, with the coarse-class label information, the model can explore the target task related fine-grained patterns more effectively. Thereafter, the classification problem can be written as
\begin{eqnarray}\label{eq:prop}
\min_{\theta} \sum_i \ell(\x_i, y_i^C)+\lambda\sum_i \ell(\x_i, y_i^I)
\end{eqnarray}
where $\lambda$ is a trade-off between the performance of the coarse-class classification and instance classification, which is corresponding to reducing intra-class variance and increasing inter-class difference, respectively. The hybrid loss functions share the same backbone for feature extraction that is denoted as $f^H(\x_i)$. The classification head is different and we let the corresponding FC layer as $W^C$ and $W^I$, respectively.

By optimizing the problem in Eqn.~\ref{eq:prop}, we prove that the performance of the learned representations can be guaranteed on the target classes as follows.
\begin{thm} \label{thm:1}
If learned representations have the bounded norm as $\forall i,j, \|f^H(\x_i)\|_2,\|\w^I_j\|_2,\|\w^C_j\|_2\leq c$
and solving the problem in Eqn.~\ref{eq:prop} such that
\[\forall i, \Pr\{y_i^I|f^H(\x_i), W^I\} \geq \alpha; \Pr\{y_i^C|f^H(\x_i), W^C\} \geq \beta\]
where $\alpha$, $\beta$ are constants that are balanced by $\lambda$, we have
\[\forall i, \quad \Pr\{y_i^F|f^H(\x_i), W^I\}\geq \alpha z h(c,\alpha,\beta)\]
where $h(c,\alpha,\beta)\leq 1$ is a constant that depends on $c$, $\alpha$, $\beta$.
\end{thm}
\paragraph{Remark}
Concretely, with the help of Eqn.~\ref{eq:ori}, we can bound the difference between examples from the same target class in $h(c,\alpha,\beta)$, while Eqn.~\ref{eq:indi} helps obtain sufficient fine-grained patterns to identify different classes for the target problem. 

It should be noted that the sub-problem of instance classification in Eqn.~\ref{eq:prop} is an $n$-class classification problem. When $n$ is large, it has to compute the scores and the corresponding gradient from $W^I\in\R^{d\times n}$ for each example, which can slow down the optimization significantly. This challenge has been extensively studied in the literature of unsupervised representation learning and mitigated by contrastive learning~\cite{abs-2002-05709,He0WXG20}. Differently, we can decompose the instance classification problem according to the coarse classes in our work, which is discussed in the following subsection.

\subsection{Large-Scale Challenge}

According to the analysis in Theorem~\ref{thm:1}, we can decompose the original problem as
\begin{eqnarray}\label{eq:large}
\min_{\theta} \sum_i \ell(\x_i, y_i^C)+\lambda\sum_{k=1}^C\sum_{i:y_i^C=k} \ell_k(\x_{i}, y_{i}^I)
\end{eqnarray}
where $\ell_k(\x_{i}, y_{i}^I)$ is the cross entropy loss defined for instance classification within the $k$-th coarse class
\begin{eqnarray*}
&&\ell_k(\x_{i}, y_{i}^I)= -\log(\Pr\{y_{i}^I|f^H(\x_i), y_{i}^C, W^I\})\\
&&=-\log(\frac{\exp(f^H(\x_i)^\top \w_{y^I_{i}}^I)}{\sum_{j:y_j^C= k} \exp(f^H(\x_i)^\top \w_j^I)})
\end{eqnarray*}
Compared with the standard instance classification, the new loss is to distinguish between the example $\x_i$ and other examples with the same coarse-class label (i.e., $y_j^C= k$) in lieu of total $n$ examples. Therefore, the computational cost of the FC layer for each example can be reduced from $\OO(dn)$ to $\OO(d n_k)$, where $n_k$ denotes the number of examples in the $k$-th coarse class.

We prove that the performance using the above speedup strategy can still be guaranteed on the target problem as stated in the following theorem.

\begin{thm}\label{thm:2}
With the same assumptions as in Theorem~\ref{thm:1}, if solving the problem in Eqn.~\ref{eq:large} such that
\begin{small}
\[\forall i, \Pr\{y_i^I|f^H(\x_i),y_i^C, W^I\}\geq \alpha; \Pr\{y_i^C|f^H(\x_i), W^C\} \geq \beta\]
\end{small}
we have
\[\forall i, \quad \Pr\{y_i^F|f^H(\x_i), W^I\}\geq \alpha' z h(c,\alpha',\beta)\]
where $\alpha' = \frac{1}{1/\alpha+(1-\beta)c''/\beta}$ and $c''$ is a constant. $h(c,\alpha',\beta)$ is a constant that depends on $c$, $\alpha'$, $\beta$.
\end{thm}
\paragraph{Remark} Compared with the guarantee in Theorem~\ref{thm:1}, the cost of relaxation is given in $\alpha'$. It contains a factor of $(1-\beta)/\beta$, which measures the performance on the coarse-class classification problem. When an example can be separated well from other coarse classes as $\beta\to 1$, the patterns obtained by solving Eqn.~\ref{eq:large} can almost recover the performance from solving the more expensive problem in Eqn.~\ref{eq:prop}.

\subsection{Instance Proxy Loss}
Till now, we theoretically analyze the behaviors of instance classification and coarse-class classification. Inspired by our analysis, we propose a novel loss to enhance the informative patterns for the target task.

A standard proxy-based triplet constraint~\cite{Qi2019} for representation learning can be written as
\[\forall \x_i,\cc_{j:j\not=y_i},\quad \|\x_i-\cc_j\|_2^2 - \|\x_i-\cc_{y_i}\|_2^2\geq \delta\]
where $\cc_j$ denotes the proxy for the $j$-th class and $\delta$ is a margin. We omit the feature extraction function $f^H(\cdot)$ for brevity. In Theorem~\ref{thm:1}, we demonstrate that the mean vector of individual classes from the same target class can be an appropriate proxy for the target task. However, the labels of target task are not available when training representations. Therefore, assuming that there are $P$ target classes, we will learn the relation with a membership variable $\mu\in\{0,1\}^{n\times P}$ ($\forall i, \sum_p \mu_{i,p}=1$) simultaneously. Specifically, we let $W^P$ denote the parameters for the $P$-class classification problem and
\begin{eqnarray}\label{eq:initw}
\w_p^P = \frac{\sum_i \mu_{i,p}\w_i^I}{\sum_i \mu_{i,p}}
\end{eqnarray}
With the proxy from averaging instance parameters, we have the triplet constraints as
\[\forall \x_i, \sum_p\frac{(1-\mu_{i,p})}{P-1}\|\x_i-\w_p^P\|_2^2 - \sum_j\mu_{i,j}\|\x_i-\w_j^P\|_2^2\geq \delta\]
To maximize the margin $\delta$, the optimization problem can be written as
\begin{small}
\begin{align}\label{eq:ploss}
\min_{\x,\mu} \sum_i\big(\sum_j\mu_{i,j}\|\x_i-\w_j^P\|_2^2 - \sum_p\frac{(1-\mu_{i,p})}{P-1}\|\x_i-\w_p^P\|_2^2\big)
\end{align}
\end{small}
The problem can be solved in an alternating manner. At each epoch, when fixing $\mu$, the representation can be optimized over $P$ classes as
\[\min_{\x} \sum_{i} \|\x_i-\w_{y_i^P}^P\|_2^2 - \sum_{p:\mu_{i,p}=0}\frac{\|\x_i-\w_p^P\|_2^2}{P-1}\]
where $\mu_{i,y_i^P} = 1$. Following the suggestion in \cite{Qi2019}, we propose an instance proxy loss to optimize the sub-problem effectively as
\begin{eqnarray}\label{eq:iploss}
\ell_{p}(\x_{i}, y_{i}^P) = -\log(\frac{\exp(f^H(\x_i)^\top \w_{y^P_{i}}^P)}{\sum_{p} \exp(f^H(\x_i)^\top \w_p^P)})
\end{eqnarray}

When fixing $\x$, the sub-problem becomes
\[\min_{\mu} \sum_i P\sum_j \mu_{i,j} \|\x_i-\w_j^P\|_2^2 - \sum_p \|\x_i-\w_p^P\|_2^2\]
Note that $W^P$ also contains $\mu$ that makes the optimization challenge. When $P$ is large, the latter term can be considered as a constant (e.g., $P=n$ for the extreme case) and the problem can be simplified as
\[\min_{\mu} \sum_i \sum_j \mu_{i,j} \|\x_i-\w_j^P\|_2^2\]
Since $W^P$ is spanned by $W^I$, we can optimize the upper-bound instead
\[\min_{\mu} \sum_i \sum_j \mu_{i,j} \|\w_i^I-\w_j^P\|_2^2+\|\x_i-\w_i^I\|_2^2\]
Without the constant term, the problem can be rewritten as
\begin{eqnarray}\label{eq:wupdate}
&&\min_{\mu,W^P} \sum_i \sum_j \mu_{i,j} \|\w_i^I-\w_j^P\|_2^2\nonumber\\
s.t.&& \w_p^P = \frac{\sum_i \mu_{i,p}\w_i^I}{\sum_i \mu_{i,p}}
\end{eqnarray}
Therefore, it becomes a standard k-means clustering problem and can be solved efficiently. To make the approximation tight, i.e., $\|\x_i-\w_i^I\|_2^2$ is small, we have to optimize the problem in Eqn.~\ref{eq:ploss} after $W^I$ is sufficiently trained.

\begin{algorithm}[t]
   \caption{Representation Learning with Coarse Labels}
   \label{alg:1}
\begin{algorithmic}
   \STATE {\bfseries Input:} training set $\{\x_i,y_i^C\}_{i=1}^n$, total epochs $T$, $M$, $P$, $\lambda_I$, $\lambda_P$
   \FOR{epoch: $t=1$ {\bfseries to} $M$}
   \STATE Optimize the problem in Eqn.~\ref{eq:large}
   \ENDFOR
   \STATE Obtain $P$ clusters with $W^I$
   \STATE Initialize $W^P$ as in Eqn.~\ref{eq:initw}
   \FOR{epoch: $t=M+1$ {\bfseries to} $T$}
   \STATE Optimize the problem in Eqn.~\ref{eq:triple}
   \STATE Update $W^P$ with fixed $W^I$ by solving Eqn.~\ref{eq:wupdate}
   \ENDFOR
\end{algorithmic}
\end{algorithm}

With the proposed instance proxy loss, the objective for representation learning becomes
\begin{eqnarray}\label{eq:triple}
&&\min_{\theta} \sum_i \ell(\x_i, y_i^C)+\lambda_I\sum_{k=1}^C\sum_{i:y_i^C=k} \ell_k(\x_{i}, y_{i}^I) \nonumber\\
&&+ \lambda_P \sum_{p=1}^P\sum_{i} \ell_{p}(\x_{i}, y_{i}^P)
\end{eqnarray}
Alg.~\ref{alg:1} summarizes the proposed algorithm. Note that the clustering can be implemented within each coarse class as suggested in Section~3.3.

\section{Experiments}
\label{sec:exp}

To evaluate the proposed method, we adopt ResNet-18~\cite{HeZRS16} as the neural network for comparison since it is the most popular deep architecture and has been widely applied for real tasks. 
We include five methods in the main comparison as follows.

\textbf{Ins}: optimize representations with instance classification only as in Eqn.~\ref{eq:indi}.

\textbf{Cos}: optimize representations with coarse-class classification only as in Eqn.~\ref{eq:ori}.

\textbf{CoIns}: learn representations with coarse-class classification and instance classification simultaneously as in Eqn.~\ref{eq:prop}.

\textbf{CoIns$_\mathrm{imp}$}: improve the efficiency by optimizing the instance classification within each coarse class as in Eqn.~\ref{eq:large}.

\textbf{Opt}: optimize representations with target labels that are not available in our problem setting. Therefore, this method provides the performance upper-bound as a reference.

ResNet-18 is trained with stochastic gradient descent (SGD). All methods in the comparison have the same backbone network and training pipeline but with different objectives and classification heads. Augmentation is important for training CNNs and we adopt both random horizontal mirroring and random crop as suggested in \cite{HeZRS16}. Other configurations on each data set follow the common practice and are elaborated in the corresponding subsections.

Three benchmark image data sets, i.e., CIFAR-100~\cite{krizhevsky2009learning}, SOP~\cite{SongXJS16}, and ImageNet~\cite{DengDSLL009}, are included for comparison. We note that all of these data sets contain both coarse-class labels and target-class labels for a comprehensive evaluation, where target-class labels are only used by ``Opt'' to provide the upper-bound of the performance.

We evaluate the performance of different representations with multiple metrics. First, we measure the accuracy on coarse classes as a side product. With more fine-grained patterns, the generalization on coarse classes can be further improved. More importantly, we evaluate the performance on the target classes by conducting the retrieval task (i.e., visual search) that motivates this work. We adopt Recall@$k$ metric as in \cite{Qi2019,SongXJS16} for comparison. The similarity for retrieval is computed by the cosine similarity using the outputs before the FC layer, i.e., $f(\x)$. \cite{Qi2019} shows that deep features learned by classification can capture the similarity between examples well.

\subsection{CIFAR-100}
In this subsection, we evaluate the methods on CIFAR-100~\cite{krizhevsky2009learning} that contains $20$ coarse classes. Each coarse class contains $5$ target classes that contribute $100$ target classes. We adopt the standard splitting, where each target class has $500$ color images for training and $100$ for test.

SGD with a mini-batch size of $256$ is applied to learn the model. Following the common practice, we set momentum to $0.9$ and weight decay as $5e^{-4}$. Each model is trained with $200$ epochs. The initial learning rate is $0.1$ and is decayed by a factor of $5$ at $\{60,120,160\}$ epochs. The $32\times 32$ images is randomly cropped from the zero-padded $40\times 40$ images for the crop augmentation. The only parameter in ``CoIns'' is $\lambda$ that balances different loss functions and we search it in $\{1,5\}\times\{10^{-i}\}_{i=0}^4$ for all experiments.

\begin{table}[ht]
\centering
\small
\begin{tabular}{|l|c|c|c|c|c|c|}
\hline
&Top1&Top5&R@1&R@2&R@4&R@8\\\hline
Ins&-&-&$22.4$&$32.9$&$46.8$&$62.6$\\\hline
Cos&85.6&97.5&$81.1$&$87.0$&$90.7$&$93.2$\\\hline
CoIns&86.3&98.2&$82.4$&$88.0$&$91.4$&$94.1$\\\hline
CoIns$_\mathrm{imp}$&$86.1$&$97.9$&$82.3$&$87.5$&$91.4$&$94.2$\\\hline
\end{tabular}
\caption{Comparison of accuracy and recall ($\%$) for $20$ coarse classes on CIFAR-100. (``-'' means NA)\label{tab:1}}
\end{table}

As a side product, Table~\ref{tab:1} summarizes both the classification and retrieval performance on the $20$ coarse classes (i.e., not the target task). First, it is surprising to observe that fine-grained patterns learned by ``CoIns'' can improve the performance on the coarse-class classification problem. It illustrates that the task-dependent patterns learned by CNNs focus on the training task and can be suboptimal for unseen examples of the same problem. Exploring more fine-grained patterns in training as suggested by ``CoIns'' can generalize the learned patterns better on unseen data. Second, ``CoIns$_\mathrm{imp}$'' has the similar performance as ``CoIns''. It is consistent with the analysis in Theorem~\ref{thm:2}. The examples with coarse-class labels can be separated well on this data set with an accuracy of more than $85\%$, which implies a large $\beta$ in Theorem~\ref{thm:2}. Therefore, the performance of ``CoIns$_\mathrm{imp}$'' can approach that of ``CoIns'' with significantly less computational cost. Note that there are $20$ coarse classes with a uniform distribution in this data set, and thus the cost of computing the fully-connected layer for instance classification in ``CoIns$_\mathrm{imp}$'' is only $5\%$ of that in ``CoIns''.

A similar observation can be obtained for the retrieval task on these 20 coarse classes. We observe that ``CoIns'' and ``CoIns$_\mathrm{imp}$'' can outperform the baseline ``Cos'' with a significant margin on R@1. ``Ins'' is included in this comparison while it provides the worst performance. It is because that the task of instance classification cannot leverage the supervised information from the coarse classes.

\begin{table}[ht]
\centering
\begin{tabular}{|l|c|c|c|c|}
\hline
&R@1&R@2&R@4&R@8\\\hline
Ins&$13.6$&$19.2$&$27.1$&$37.3$\\\hline
Cos&$37.1$&$51.6$&$67.0$&$79.9$\\\hline
CoIns&$57.0$&$68.0$&$77.5$&$85.5$\\\hline
CoIns$_\mathrm{imp}$&$56.6$&$68.0$&$77.5$&$85.1$\\\hline
CoIns$^*$&$60.8$&$71.2$&$79.2$&$85.5$\\\hline
CoIns$^{**}$&$60.5$&$71.1$&$79.8$&$86.5$\\\hline
CoInsP$^{**}$&$62.0$&$71.7$&$80.2$&$86.6$\\\hline
Opt&$71.8$&$78.8$&$84.1$&$88.3$\\\hline
\end{tabular}
\caption{Comparison of recall ($\%$) for $100$ classes on CIFAR-100. CoIns$^*$ adopts cosine softmax while CoIns$^{**}$ has both cosine softmax and MLP head as in \cite{abs-2006-14618}.\label{tab:2}}
\end{table}

More importantly, the comparison on the target retrieval task of $100$ classes is demonstrated in Table~\ref{tab:2}. Evidently, both ``Cos'' and ``Ins'' cannot handle the retrieval task well. As illustrated in our analysis, ``Cos'' lacks the fine-grained patterns, i.e., small inter-class difference and ``Ins'' lacks the guidance to filter massive noisy patterns, i.e., large intra-class variance. By complementing each other in ``CoIns'', the performance can be dramatically improved. The R@1 of ``CoIns'' is better than ``Cos'' by about $20\%$ and surpasses ``Ins'' by more than $40\%$. It confirms the observation in Theorem~\ref{thm:1} that the proposed method can explore the fine-grained patterns sufficiently and effectively for the target task when only coarse-class labels are available. Without doubt, ``Opt'' provides the best performance when target-class labels are available for training. Compared to ``Opt'', we can observe that R@4 of ``CoIns'' is better than R@1 of ``Opt'' and is comparable to R@2 of ``Opt''. It means that when only coarse-class labels are available, by optimizing the objective in Eqn.~\ref{eq:prop}, the learned model can handle the target retrieval task well by retrieving two additional examples. Finally, the negligible difference between the performance of ``CoIns'' and ``CoIns$_\mathrm{imp}$'' implies that ``CoIns$_\mathrm{imp}$'' is efficiently applicable for real-world applications.

Many recent works including SimCLR~\cite{Ting5709}, MoCo-v2~\cite{abs-2003-04297}, and PIC~\cite{abs-2006-14618} indicate that some additional components are essential for the success of unsupervised learning on ImageNet. Therefore, we introduce these components to ``CoIns'' to evaluate their effects in our problem. Specifically, three components including cosine softmax, MLP, and strong augmentation, are compared. We observe that strong augmentation always hurts the performance. It may be due to the fact that strong augmentation introduces too much noise for CIFAR, so we ignore its results in Table~\ref{tab:2}. ``CoIns'' with cosine softmax and with both cosine softmax and MLP are referred as \textbf{CoIns$^*$} and \textbf{CoIns$^{**}$}, respectively.

From Table~\ref{tab:2}, it is evident that ``CoIns$^*$'' can further improve the performance of ``CoIns'' with a significant margin of $3\%$, which is consistent with the observation in \cite{abs-2006-14618}. Since applying unit norm for both representations of examples and parameters in the FC layer, it can have a better guarantee as illustrated in Theorem~\ref{thm:1}. However, ``CoIns$^{**}$'' with an additional MLP head cannot surpass ``CoIns$^*$'' when retrieved examples are limited. The reason may be from the low resolution of images in CIFAR. Finally, we incorporate these two components into ``CoInsP'' that adds the proposed instance proxy (IP) loss for training as in Eqn.~\ref{eq:iploss}. In the experiment, we add the IP loss to ``CoIns'' after half of the training process, i.e., $100$ epochs. To make the approximation tight as analyzed in Section~3.4, we have a large $P$ as $P=10,000$. By mimicking the target classes and enhancing instance classification, ``CoInsP$^{**}$'' achieves the best performance that is closest to ``Opt''. Moreover, R@2 of ``CoInsP$^{**}$'' already has the similar performance to R@1 of ``Opt'', which demonstrates the effectiveness of the proposed method.

\subsection{Stanford Online Products}
Then, we evaluate different algorithms in a challenging online shopping scenario. Stanford Online Products (SOP)~\cite{SongXJS16} collects $120,053$ product images from eBay.com. There are a total of $22,634$ classes from $12$ coarse classes. Therefore, each target class contains very limited number of examples. Since there is no public splitting on this data set for classification, we randomly sample $80,000$ images for training and the rest for test. We then filter all classes that contain only a single example in the test set. This leads to $13,160$ target classes for evaluation.

For training, we adopt the suggested configuration as in \cite{HeZRS16}. Specifically, the model is learned from scratch with $90$ epochs. The initial learning rate is $0.1$ and decayed by a factor of $10$ at $\{30,60\}$ epochs. The similar results as CIFAR for coarse-class classification and retrieval can be found in appendix.

\begin{table}[ht]
\centering
\begin{tabular}{|c|c|c|c|}
\hline
&R@1&R@10&R@100\\\hline
Ins&$25.5$&$38.3$&$54.9$\\\hline
Cos&$21.8$&$34.4$&$52.7$\\\hline
CoIns&$35.8$&$51.8$&$69.3$\\\hline
CoIns$_\mathrm{imp}$&$35.3$&$50.5$&$67.4$\\\hline
CoIns$^*$&$38.1$&$54.2$&$70.6$\\\hline
CoIns$^{**}$&$42.7$&$58.2$&$73.7$\\\hline
CoInsP$^{**}$&$43.5$&$59.0$&$74.3$\\\hline
Opt&$46.5$&$61.6$&$75.2$\\\hline
\end{tabular}
\caption{Comparison of recall ($\%$) for $13,160$ classes on SOP. CoIns$^*$ adopts cosine softmax while CoIns$^{**}$ has both cosine softmax and MLP head as in \cite{abs-2006-14618}. \label{tab:6}}
\end{table}

The retrieval performance on the target classes is shown in Table~\ref{tab:6}. Considering the well-known difficulty of this task, we report the Recall@\{1,10,100\} as suggested in \cite{Qi2019,SongXJS16}. First, we can observe that ``CoIns'' outperforms ``Cos'' by $14\%$ on R@1. It demonstrates that our method can be applied for online shopping scenario when there is limited supervision. Besides, even with the supervised information on target classes, R@1 of ``Opt'' is less than $50\%$, which shows that retrieval in online shopping is an important but challenging application. With more retrieved examples, recall of ``CoIns'' can outperform $50\%$ as shown in R@10. Note that customers for online shopping tend to review only the top ranked items, which is known as position bias~\cite{JoachimsGPHG17}. Therefore, improving R@10 is important for better customer experience.

With the additional components, ``CoIns$^*$'' surpasses ``CoIns'', while ``CoIns$^{**}$'' shows an even better performance that is closer to ``Opt''. It demonstrates that cosine softmax can benefit both of low-resolution images and high-resolution ones, while MLP is especially effective for high-resolution images as in SOP. Finally, ``CoInsP$^{**}$'' with an additional loss after $45$ epochs demonstrates the best performance among variants of ``CoIns''. It shows that the informative patterns can be further captured using the proposed instance proxy loss.

\begin{figure}[ht]
\centering
\includegraphics[height=1.6in]{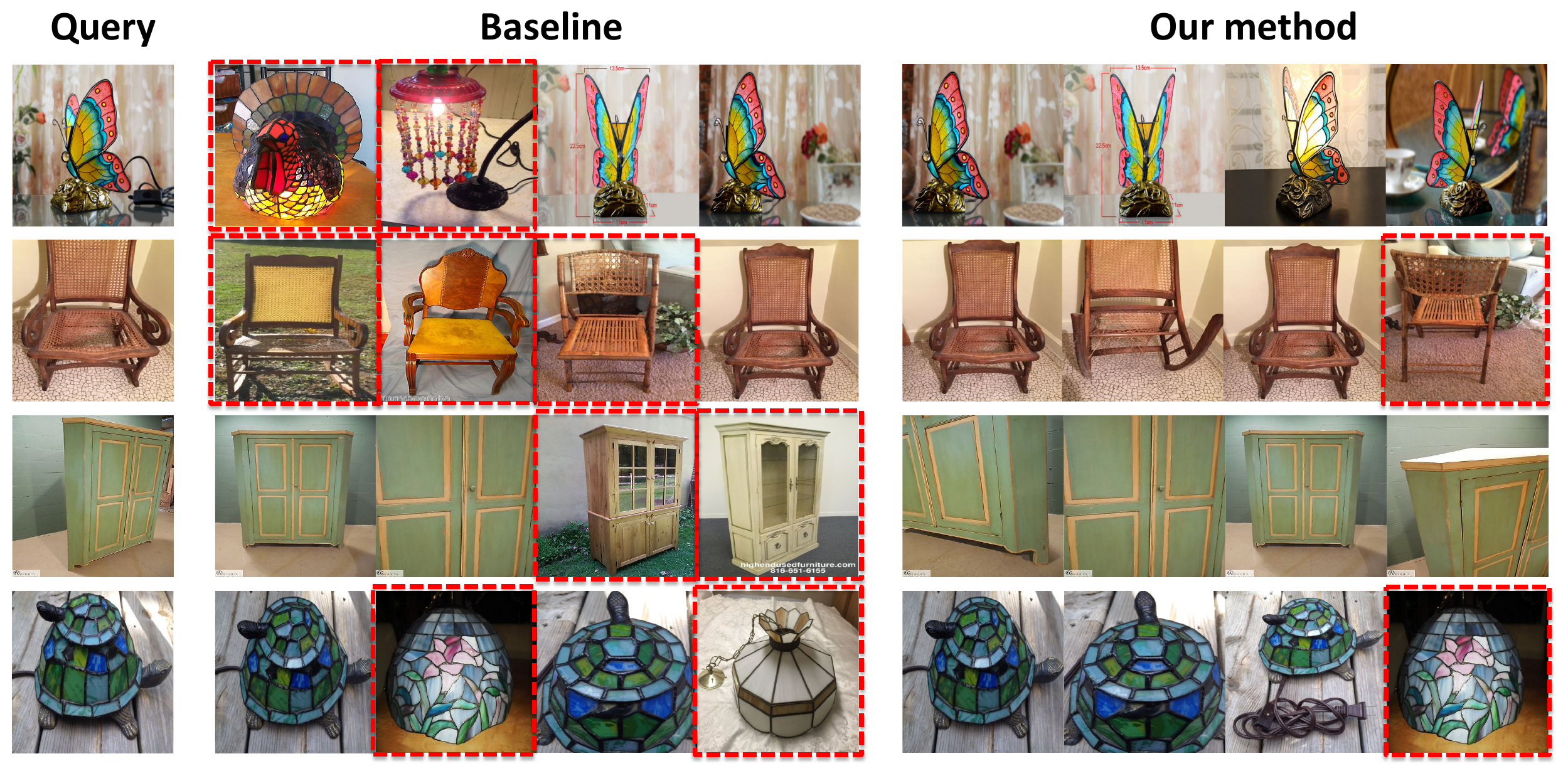}
\caption{Examples of retrieved images from \textbf{Cos} (i.e., baseline) and \textbf{CoIns} (i.e., our method) on SOP. The examples from a different target class are denoted with red bounding-boxes.\label{fig:4}}
\end{figure}

We illustrate the retrieved images on SOP in Fig.~\ref{fig:4}. Evidently, there are many similar products from different target classes in online shopping, which makes the application very challenging. Given a query image, it is hard for ``Cos'' (i.e., baseline) to retrieve appropriate similar items. By learning fine-grained patterns sufficiently as in ``CoIns'', the examples from different target classes are eliminated from the top ranked items.

\subsection{ImageNet}
Finally, we compare different methods on ImageNet~\cite{DengDSLL009}. ImageNet is a popular benchmark data set for visual categorization. It contains $1,000$ classes and each class has about $1,200$ images. These classes are organized according to WordNet~\cite{fellbaum1998} and there can be $11$ coarse classes in ImageNet as analyzed in \cite{QianHL20}. Each coarse class can have multiple target classes. For example, the coarse class ``dog'' has $118$ different species of dogs and ``bird'' contains $59$ different species of birds.

Instance classification has been extensively studied on ImageNet and many sophisticated algorithms have been developed. To make the comparison fair, we adopt one state-of-the-art method, MoCo-v2~\cite{abs-2003-04297}, as the substitute of instance classification in Eqn.~\ref{eq:prop}. We implement our method by adding coarse-class classification to the official code of MoCo. The training follows the configuration of MoCo-v2 with $200$ epochs. We also adopt ResNet-50 rather than ResNet-18 in the comparison to align with the result of MoCo that is only implemented with ResNet-50. 

\begin{table}[ht]
\centering
\small
\begin{tabular}{|l|c|c|c|c|c|c|}
\hline
&Top1&Top5&R@1&R@2&R@4&R@8\\\hline
MoCo-v2&$67.5$&$88.0$&$42.8$&$52.9$&$62.4$&$71.1$\\\hline
Cos&$60.4$&$83.0$&$21.1$&$28.9$&$37.9$&$48.2$\\\hline
CoIns&$70.4$&$89.9$&$51.4$&$61.5$&$70.7$&$78.4$\\\hline
Opt&$76.2$&$92.9$&$66.4$&$75.3$&$82.1$&$87.3$\\\hline
\end{tabular}
\caption{Comparison of accuracy and recall ($\%$) for $1,000$ classes on ImageNet.\label{tab:7}}
\end{table}

Table~\ref{tab:7} compares different methods on ImageNet. The performance of MoCo-v2 is directly borrowed from the official pre-trained model while that of ``Opt'' is from the pre-trained model provided by PyTorch\footnote{https://pytorch.org/vision/stable/models.html}. First, we can observe that MoCo-v2 is worse than ``Opt'' by more than $20\%$ on R@1. It demonstrates that without labels, instance classification cannot learn the patterns well related to the target classes. However, the performance ``Cos'' is even worse since we only introduce $11$ coarse classes that cannot handle the inter-class difference for the target task. By incorporating these coarse classes as in our method, R@1 using learned representations can be increased from $42.8\%$ to $51.4\%$. It confirms our analysis in Theorem~\ref{thm:1} that coarse classes help to eliminate noisy patterns and can improve the performance on the target task. 

Besides, we also include the comparison of classification on $1,000$ target classes in Table~\ref{tab:7}. The performance is evaluated by a linear classifier with fixed representations and the classifier is learned by the standard pipeline provided in MoCo-v2. Note that target labels will be applied for training linear classification. First, the accuracy of MoCo achieves $67.5\%$ after fine-tuning with target-class labels. It shows that unsupervised instance classification relies on target label information to filter noisy patterns in representations. With more related patterns in ``CoIns'', the Top1 accuracy can achieve $70.4\%$, which is about $3\%$ better than MoCo-v2. This further demonstrates our proposed method. It also implies that even with full supervised information from the target task for fine-tuning, the representations learned from instance classification is worse than our proposal.

\section{Conclusion}
\label{sec:conclude}
In this work, we propose an algorithm to explore fine-grained patterns sufficiently with an access of only coarse-class labels for training. The empirical study on benchmark data sets confirms the effectiveness of our proposed method and its theoretical guarantee. Besides, we propose a new instance proxy loss to further improve the performance according to our theoretical analysis. 

Considering that the number of unlabeled data is significantly larger than that of labeled data, incorporating unlabeled data to improve the performance can be our future work. Moreover, there can be various weakly supervised information besides labels (e.g., triplet constraints, multiple views), exploring and incorporating more coarse information to catch up the performance upper-bound is also an interesting future direction.

{\small
\bibliographystyle{ieee_fullname}
\bibliography{fgp}
}

\appendix

\section{Theoretical Analysis}
\label{sec:theorem}

\subsection{Proof of Lemma~1}
\begin{proof}

To simplify the proof, we assume that each target class contains $z$ examples as $zF = n$.
Then, we define
\begin{eqnarray*}
&&\Pr\{y_i^F|f(\x_i),W^I\} = \frac{\exp(f(\x_i)^\top \bar{\w}_{y_i}^I)}{\sum_s^F \exp(f(\x_i)^\top \bar{\w}_s^I)}
\end{eqnarray*}
where $\bar{\w}_s^I = \frac{1}{z}\sum_{y_j^F= s}\w_j^I$ averaging over the parameters from the same target class. According to the Jensen's inequality, we have
\[\exp(f(\x_i)^\top \bar{\w}_s^I)\leq \frac{1}{z}\sum_{y_j^F=s} \exp(f(\x_i)^\top \w_j^I)\]
Therefore, we have
\begin{align}\label{eq:prop2}
&\Pr\{y_i^F|f(\x_i),W^I\} \nonumber\\
&\geq z\exp(f(\x_i)^\top \bar{\w}_{y_i}^I - f(\x_i)^\top\w_{y_i}^I)\Pr\{y_i^I|\x_i,W^I\}\nonumber\\
&\geq z\alpha \exp(f(\x_i)^\top \bar{\w}_{y_i}^I - f(\x_i)^\top\w_{y_i}^I)
\end{align}
where $\exp(f(\x_i)^\top \bar{\w}_{y_i}^I - f(\x_i)^\top\w_{y_i}^I)$ measures the distance from an individual example to other examples from the same target class. It cannot be bounded well by only solving the problem of instance classification.

\subsection{Proof of Theorem~1}

First, by optimizing the proposed classification problem, we assume
\[\forall i,\quad \Pr\{y_i^I|f(\x_i), W^I\} = \frac{\exp(f(\x_i)^\top \w_{y_i}^I)}{\sum_j^n \exp(f(\x_i)^\top \w_j^I)}\geq \alpha\]
and
\[\forall i,\quad \Pr\{y_i^C|f(\x_i), W^C\} = \frac{\exp(f(\x_i)^\top \w_{y_i}^C)}{\sum_j^C \exp(f(\x_i)^\top \w_j^C)}\geq \beta\]
By assuming the residual is lower bounded by constants $a$ and $b$ as
\begin{small}
\[\forall i, \sum_{j,j\not=y_i^I}^n \exp(f(\x_i)^\top \w_j^I)\geq a; \sum_{j,j\not=y_i^C}^C \exp(f(\x_i)^\top \w_j^C)\geq b\]
\end{small}
we have
\begin{small}
\[\exp(f(\x_i)^\top \w_{y_i}^I)\geq \frac{\alpha}{1-\alpha}(\sum_{j,j\not=y_i^I}^n \exp(f(\x_i)^\top \w_j^I))\geq \frac{a\alpha}{1-\alpha}\]
\end{small}
and
\begin{small}
\[\exp(f(\x_i)^\top \w_{y_i}^C)\geq \frac{\beta}{1-\beta}(\sum_{j,j\not=y_i^C}^C \exp(f(\x_i)^\top \w_j^C))\geq \frac{b\beta}{1-\beta}\]
\end{small}
which leads to
\begin{small}
\[\forall i,\quad f(\x_i)^\top \w_{y_i}^I\geq \log(\frac{a\alpha}{1-\alpha});\quad f(\x_i)^\top \w_{y_i}^C\geq \log(\frac{b\beta}{1-\beta})\]
\end{small}

To guarantee the performance on the target classification problem, we have to bound $\exp(f(\x_i)^\top \bar{\w}_{y_i}^I - f(\x_i)^\top\w_{y_i}^I)$ as illustrated in Eqn.~\ref{eq:prop2}. Now, it can be bounded with the help from solving the coarse-class classification problem. Specifically, the instance similarity can be bounded as
\begin{align*}
&f(\x_i)^\top\w_j^I - f(\x_i)^\top\w_{y_i}^I \\
&= f(\x_i)^\top f(\x_j)+f(\x_i)^\top(\w_j^I - f(\x_j))\\
&-f(\x_i)^\top f(\x_i)+f(\x_i)^\top(f(\x_i)^\top - \w_{y_i}^I)
\end{align*}

Then, we can bound each term as follows. First, the distance between an example to its individual label representation (i.e., $\w$) can be bounded by solving the individual classification problem as
\begin{small}
\begin{align*}
&f(\x_i)^\top(\w_j^I - f(\x_j)) \geq -\|f(\x_i)^\top(\w_j^I - f(\x_j))\|_2\\
&\geq -\|f(\x_i)\|_2\|\w_j^I - f(\x_j)\|_2 (\mbox{Cauchy-Schwarz inequality})\\
&\geq -c\|\w_j^I - f(\x_j)\|_2\\
&\geq -c\sqrt{2c^2-2\log(a\alpha/(1-\alpha))}
\end{align*}
\end{small}
With the similar analysis, we have
\[f(\x_i)^\top(f(\x_i)^\top - \w_{y_i}^I)\geq -c\sqrt{2c^2-2\log(a\alpha/(1-\alpha))}\]

Note that examples from the same target class also share the same coarse-class label. Therefore, the distances between examples from the same target class can be bounded as
\begin{align*}
&f(\x_i)^\top f(\x_j)-f(\x_i)^\top f(\x_i)\geq -c\|f(\x_j) - f(\x_i)\|_2\\
&\geq -c(\|f(\x_j) - \w_{y_i}^C\|_2+\|f(\x_i) - \w_{y_i}^C\|_2)\\
&\geq -2c\sqrt{2c^2 - 2\log(b\beta/(1-\beta))}
\end{align*}
Combining them together, we have
\begin{align*}
&\exp(f(\x_i)^\top \bar{\w}_{y_i}^I - f(\x_i)^\top\w_{y_i}^I)\\
&\geq \exp\Big(-\frac{2c(z-1)}{z}\big(\sqrt{2c^2-2\log(a\alpha/(1-\alpha))}\\
&+\sqrt{2c^2-2\log(b\beta/(1-\beta))}\big)\Big)
\end{align*}
Taking it back to Eqn.~\ref{eq:prop2}, we can observe the desired result
\[\Pr\{y_i^F|f(\x_i),W^I\} \geq \alpha z h(c,\alpha,\beta)\]
where
\begin{small}
\begin{align*}
&h(c,\alpha,\beta) =  \exp\Big(-\frac{2c(z-1)}{z}\\
&\big(\sqrt{2c^2-2\log(a\alpha/(1-\alpha))}+\sqrt{2c^2-2\log(b\beta/(1-\beta))}\big)\Big)
\end{align*}
\end{small}
\end{proof}

\subsection{Proof of Theorem~2}
\begin{proof}
Following the analysis in Theorem~1, we assume
\[\forall i, \Pr\{y_i^I|\x_i,y_i^C, W^I\} = \frac{\exp(f(\x_i)^\top \w_{y_i}^I)}{\sum_{j:j=y_i^C} \exp(f(\x_i)^\top \w_j^I)}\geq \alpha\]
\[\forall i, \Pr\{y_i^C|\x_i, W^C\} = \frac{\exp(f(\x_i)^\top \w_{y_i}^C)}{\sum_j^C \exp(f(\x_i)^\top \w_j^C)}\geq \beta\]
\begin{align*}
&\forall i,\quad \sum_{j=y_i^C,j\not=y_i^I} \exp(f(\x_i)^\top \w_j^I)\geq a\\
& \forall i,\quad \sum_{j\not=y_i^C}^C \exp(f(\x_i)^\top \w_j^C)\geq b
\end{align*}

Compared with the analysis for Theorem~1, if we can bound $\Pr\{y_i^I|\x_i,W^I\}$, the performance on the target classification task can be guaranteed. First, we try to bound the similarity between the example and the individual class. Considering $\forall j, j\not=y_i^C$, we have
\begin{align*}
&f(\x_i)^\top \w_j^I = f(\x_i)^\top f(\x_j)+f(\x_i)^\top (\w_j^I - f(\x_j))\\
&\leq f(\x_i)^\top f(\x_j) +c\|\w_j^I - f(\x_j)\|_2\\
&\leq f(\x_i)^\top \w_j^C+c (\|f(\x_j)-\w_j^C\|_2+\|\w_j^I - f(\x_j)\|_2)
\end{align*}
Note that
\[(1-\beta)\exp(f(\x_i)^\top \w_{y_i}^C)\geq \beta \sum_{j\not=y_i^C}^C \exp(f(\x_i)^\top \w_j^C)\]
we have
\[\forall j\not=y_i^C,\quad f(\x_i)^\top \w_j^C\leq \log(\frac{1-\beta}{\beta}) + f(\x_i)^\top \w_{y_i}^C\]
Therefore, the similarity can be further bounded as
\begin{eqnarray*}
&&f(\x_i)^\top \w_j^I \leq \log(\frac{1-\beta}{\beta}) + f(\x_i)^\top \w_{y_i}^C\\
&&+c (\|f(\x_j)-\w_j^C\|_2+\|\w_j^I - f(\x_j)\|_2)\\
&&\leq \log(\frac{1-\beta}{\beta}) + f(\x_i)^\top (\w_{y_i}^C-\w_{y_i}^I+\w_{y_i}^I) \\
&&+c(\|f(\x_j)-\w_j^C\|_2+\|\w_j^I - f(\x_j)\|_2)\\
&&\leq \log(\frac{1-\beta}{\beta})+ f(\x_i)^\top \w_{y_i}^I\\
&&+c(\|\w_{y_i}^C-f(\x_i)\|_2+\|f(\x_i) - \w_{y_i}^I\|_2\\
&&+\|f(\x_j)-\w_j^C\|_2+\|\w_j^I - f(\x_j)\|_2)
\end{eqnarray*}
Note that the distance between an example and its corresponding parameters $\w_i$ can be bounded as in Theorem~1. Therefore, we have
\begin{small}
\[\forall i, j:j\not=y_i^C, \exp (f(\x_i)^\top \w_j^I)\leq \frac{1-\beta}{\beta}c'\exp(f(\x_i)^\top \w_{y_i}^I)\]
\end{small}
where 
\begin{eqnarray*}
&&c' = \exp(2c(\sqrt{2c^2-2\log(a\alpha/(1-\alpha))}\\
&&+\sqrt{2c^2-2\log(b\beta/(1-\beta))}))
\end{eqnarray*}

Then, we have
\begin{eqnarray*}
&&\Pr\{y_i^I|\x_i, W^I\} = \frac{\exp(f(\x_i)^\top \w_{y_i}^I)}{\sum_j^n \exp(f(\x_i)^\top \w_j^I)}\\
&&=\frac{1}{\frac{\sum_{j:j= y_i^C} \exp(f(\x_i)^\top \w_j^I)}{\exp(f(\x_i)^\top \w_{y_i}^I)}+\frac{\sum_{j:j\not= y_i^C} \exp(f(\x_i)^\top \w_j^I)}{\exp(f(\x_i)^\top \w_{y_i}^I)}}\\
&& \geq \frac{1}{1/\alpha+(1-\beta)c'M/\beta}
\end{eqnarray*}
where $M$ denotes the quantity of examples from different coarse classes as $M = |\{\x_j: y_j^C\not=y_i^C\}|$. Letting 
\[\alpha' = \frac{1}{1/\alpha+(1-\beta)c''/\beta}\]
where $c'' = c'M$, we can obtain the guarantee by the similar analysis as in Theorem~1.
\end{proof}

\section{Experiments}

\subsection{Synthetic Data}

\begin{figure}[!ht]
\centering
\includegraphics[height=1.1in]{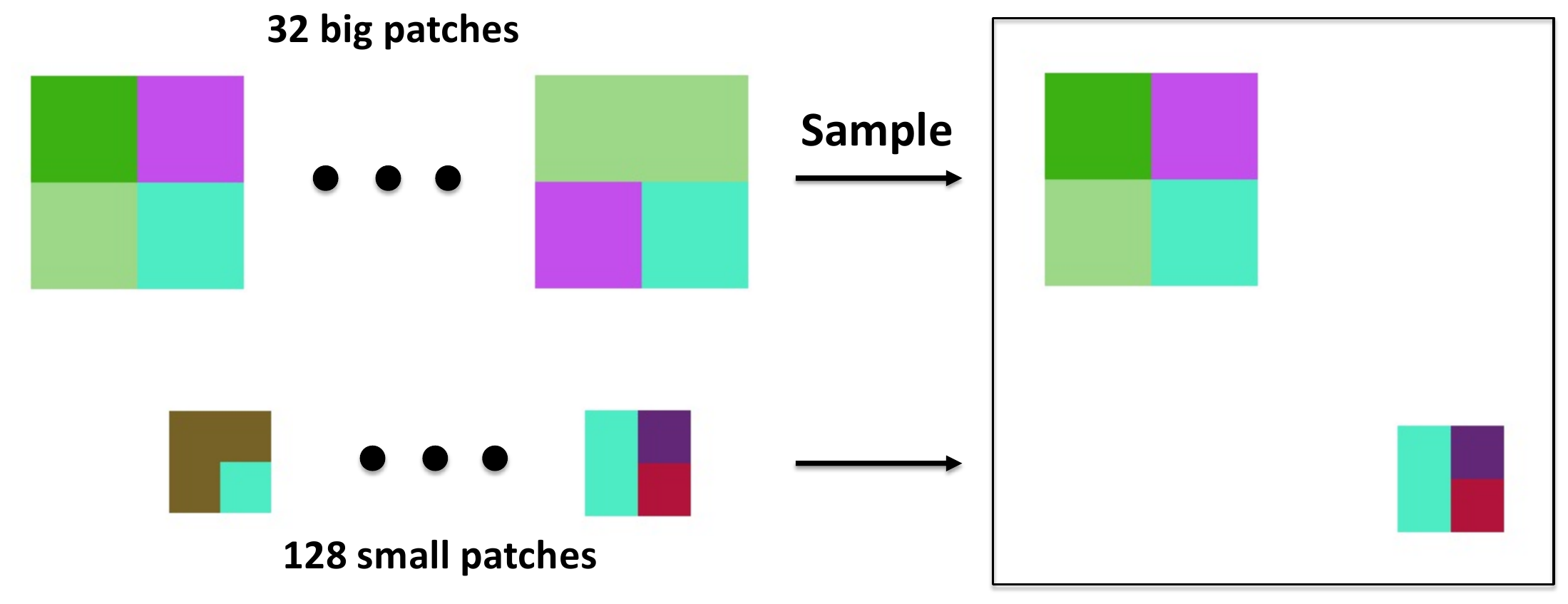}
\caption{Data generation procedure for synthetic data.\label{fig:6}}
\end{figure}

Besides comparing the performance on real-world data sets, we conduct an experiment on the synthetic data to illustrate the difference between patterns learned using different training labels on the same data set. The synthetic data is generated as follows. First, we randomly generate $32$ big color patches and $128$ small color patches as a pool of patches. Given a blank image, a big patch and a small patch are randomly sampled from the pool, and then added to the image. Finally, $512$ images are obtained. Fig.~\ref{fig:6} illustrates the process.

For the synthetic data, coarse classes are defined with big patches and target classes are defined by small patches. Consequently, there are $32$ coarse classes and $128$ target classes in the data set. To investigate the different patterns learned by the neural network, we train the model with the objective of ``Ins'', ``Cos'' and ``Opt'', respectively. After that, we visualize the spatial attention maps of different models to illustrate the patterns exploited by these models. The detailed algorithm for computing attention maps can be found in~\cite{ZagoruykoK17}.

Fig.~\ref{fig:2} shows the attention maps of these three tasks. First, we can observe that deep learning can capture the most discriminative parts for a given task. For example, it can identify the big patches for the $32$-class classification task and the small patches for the $128$-class classification task. Second, the learned patterns for them are totally different. When training the model with the objective of ``Cos'', the neural network will ignore small patches, which are essential for the 128-class classification problem. It demonstrates that the patterns learned from the conventional pipeline with coarse labels only can be inappropriate for the target task. Finally, optimizing the loss for identifying each example as ``Ins'' can explore all patterns in images which may introduce additional noise for the target task (i.e., 128-class).

\subsection{Stanford Online Products}
Results for coarse-class classification and retrieval are summarized in Table~\ref{tab:5}. It is evident that even when the target task is consistent with the training labels, exploring fine-grained patterns can achieve additional gain.

\begin{table}[ht]
\centering
\small
\begin{tabular}{|l|c|c|c|c|c|c|}
\hline
&Top1&Top5&R@1&R@2&R@4&R@8\\\hline
Cos&$80.1$&$97.3$&$76.6$&$84.3$&$89.5$&$93.2$\\\hline
CoIns$_\mathrm{imp}$&$80.6$&$97.5$&$77.1$&$84.5$&$89.7$&$93.4$\\\hline
\end{tabular}
\caption{Comparison of accuracy and recall ($\%$) for $12$ coarse classes on SOP.\label{tab:5}}
\end{table}

\subsection{Ablation Study for Instance Proxy Loss}
In this subsection, we conduct experiments to evaluate the effect of instance proxy loss in ``CoInsP$^{**}$''. CIFAR-100 is adopted for the ablation study. The weight for the corresponding loss function is set as $\lambda_P=1$.
\paragraph{Effect of $M$} First, we investigate the effect of the number of epochs before the instance proxy loss is added for optimization. Since the learning rate will be first decayed at the $60$-th epoch, we add the loss at $\{60,80,100,120\}$ epochs and summarize the results in Table~\ref{tab:pcm}. The performance of ``CoIns$^{**}$'' is included as a baseline. Apparently, including the instance proxy loss can improve the performance after sufficient training. It is consistent with our analysis that the parameters from instance classification can be applied to generate appropriate proxies for the target task when those parameters can identify individual examples well. However, if we have the loss after another decay of learning rate at the $120$-th epoch, the improvement vanishes. This phenomenon is due to the fact that the learning rate is too small to exploit the additional informative patterns effectively..

\begin{table}[ht]
\centering
\begin{tabular}{|l|c|c|c|c|}
\hline
$M$&R@1&R@2&R@4&R@8\\\hline
Baseline&60.5&71.1&79.8&86.5\\\hline
60&61.4&71.5&79.7& 86.3\\\hline
80&61.7&71.7&80.0&86.5 \\\hline
100&\textbf{62.0}&\textbf{71.7}&\textbf{80.2}&\textbf{86.6}\\\hline
120&60.8&70.6&79.6&86.5\\\hline
\end{tabular}
\caption{Ablation study on $M$ before the instance proxy loss is added.\label{tab:pcm}}
\end{table}

\paragraph{Effect of $P$}
Table~\ref{tab:pcq} compares the performance by varying $P$. When $P=25,000$, the number of clusters is half of that of total examples. There is no sufficient information within each cluster for CIFAR-100, since each cluster contains only about 2 instances. In contrast, a small $P$ will lead to more aggregated clusters and can contain patterns that are related to the target task. If $P$ is too small (i.e., the cluster size is too large), additional noise can be introduced and result in the suboptimal results as illustrated when $P=500$. It confirms our analysis that a large $P$ is important for the tight approximation. 

\begin{table}[ht]
\centering
\begin{tabular}{|l|c|c|c|c|}
\hline
$P$&R@1&R@2&R@4&R@8\\\hline
25,000&60.5& 70.8&  79.6&  86.6\\\hline
10,000&\textbf{62.0}&\textbf{71.7}&\textbf{80.2}&\textbf{86.6}\\\hline
5,000&61.2&71.2&78.8&85.4 \\\hline
1,000&58.9&69.0&76.5&82.3\\\hline
500&57.9&67.9&76.0&82.1\\\hline
\end{tabular}
\caption{Ablation study on the number of clusters when using the instance proxy loss.\label{tab:pcq}}
\end{table}

\end{document}